\documentclass[runningheads]{llncs}
\usepackage{graphicx}
\usepackage{tikz}
\usepackage{amsmath,amssymb} 
\usepackage{color}
\usepackage{amsmath}
\usepackage{amssymb}
\usepackage{booktabs}
\usepackage{multirow}
\usepackage{rotating}
\usepackage{float}
\usepackage{xcolor}
\usepackage{textcomp}
\usepackage{xspace}
\usepackage[pagebackref,breaklinks,colorlinks]{hyperref}
\usepackage{algorithm}
\usepackage[noend]{algpseudocode}
\usepackage{wrapfig}

\usepackage[capitalize]{cleveref}

\begin{document}
\pagestyle{headings}
\sloppy
\mainmatter
\input{defs}
\title{Motion Inspired Unsupervised Perception and Prediction in Autonomous Driving}

\titlerunning{Motion Inspired Unsupervised Perception and Prediction}

\author{Mahyar Najibi\thanks{Equal contribution} \and
Jingwei Ji\samethanks \and
Yin Zhou\samethanks\thanks{Corresponding author} \and Charles R. Qi \and
Xinchen Yan \and\\
Scott Ettinger \and Dragomir Anguelov}
\authorrunning{M. Najibi et al.}

\institute{Waymo LLC\\
\email{\{najibi,jingweij,yinzhou,rqi,xcyan,settinger,dragomir\}@waymo.com}}
\maketitle

\begin{abstract}
Learning-based perception and prediction modules in modern autonomous driving systems typically rely on expensive human annotation and are designed to perceive only a handful of predefined object categories. This closed-set paradigm is insufficient for the safety-critical autonomous driving task, where the autonomous vehicle needs to process arbitrarily many types of traffic participants and their motion behaviors in a highly dynamic world. To address this difficulty, this paper pioneers a novel and challenging direction, \textit{i.e.,} training perception and prediction models to understand \emph{open-set} moving objects, with no human supervision. Our proposed framework uses self-learned flow to trigger an automated meta labeling pipeline to achieve automatic supervision. 
3D detection experiments on the Waymo Open Dataset show that our method significantly outperforms classical unsupervised approaches and is even competitive to the counterpart with supervised scene flow. We further show that our approach generates highly promising results in open-set 3D detection and trajectory prediction, confirming its potential in closing the safety gap of fully supervised systems.  

\end{abstract}

\cutsectionup
\section{Introduction}
\label{intro}
\cutsectiondown

Modern 3D object detection~\cite{shi2019pointrcnn,qi2019deep,REF:Yang_3DSSD_2020_CVPR,zhou2018voxelnet} and trajectory prediction models~\cite{multipath_2019,DenseTNT_Gu_2021_ICCV,FaF_Luo_2018_CVPR,ye2021tpcn} are often designed to handle a predefined set of object types and rely on costly human annotated datasets for their training. While such paradigm has achieved great success in pushing the capability of autonomy systems, it has difficulty in generalizing to the \emph{open-set} environment that includes a long-tail distribution of object types far beyond the predefined taxonomy. Towards solving the 3D object detection and behavior prediction of those open-set objects, an alternative and potentially more scalable approach to supervised training is unsupervised perception and prediction.

One central problem in autonomous driving is perceiving the amodal shape of moving objects in space and forecasting their future trajectories, such that the planner and control systems can maneuver safely. 
As motion estimation (also known as the scene flow problem) is a fundamental task agnostic to the scene semantics~\cite{Self_pillarmotion_Luo_2021_CVPR},
it provides an opportunity to address the problem of perception and prediction of open-set moving objects, without any human labels. This leads to our motion-inspired unsupervised perception and prediction system.

Using only LiDAR, our system decomposes the unsupervised, open-set learning task to two steps, as shown in \cref{fig:teaser}: (1) Auto Meta Labeling (AML) assisted by scene flow estimation and temporal aggregation, which generates pseudo labels of any moving objects in the scene; (2) Training detection and trajectory prediction models based on the auto meta labels. Realizing such an automatic supervision, we transform the challenging open-set learning task to a known, well-studied task of supervised detection or behavior prediction model training.

To derive high-quality auto meta labels, we propose two key technologies: an unsupervised scene flow estimation model and a flow-based object proposal and concept construction approach.
Most prior works on unsupervised scene flow estimation~\cite{wu2020pointpwc,li2021_nsf,mittal2020just,graph_prior_2020} optimize for the overall flow quality without specifically focusing on the moving objects or considering the usage of scene flow for onboard perception and prediction tasks. 
For example, the recently proposed Neural Scene Flow Prior (NSFP)~\cite{li2021_nsf} achieved state-of-the-art performance in overall scene flow metrics by learning to estimate scene flow through run-time optimization, without any labels. However, there are too many false positive flows generated for the background, which makes it not directly useful for flow-based object discovery.
To tackle its limitations, we extend NSFP to a novel, more accurate and scalable version termed NSFP++. Based on the estimated flow, we propose an automatic pipeline to generate proposals for all moving objects and reconstruct the object shapes (represented as amodal 3D bounding boxes) through tracking, shape registration and refinement. The end product of the process is a set of 3D bounding boxes and tracklets.
Given the auto labels, we can train top-performing 3D detection models to localize the open-set moving objects and train behavior prediction models to forecast their trajectories. 

Evaluated on the Waymo Open Dataset~\cite{sun2019scalability}, we show that our unsupervised and data-driven method significantly outperforms non-parametric clustering based approaches and is even competitive to supervised counterparts (using ground truth scene flow). More importantly, our method substantially extends the capability of handling open-set moving objects for 3D detection and trajectory prediction models, leading to a safety improved autonomy system. 
\cutsectionup
\section{Related Works}
\label{related_works}
\cutsectiondown

\subsubsection{LiDAR-based 3D Object Detection.}
Fully supervised 3D detection based on point clouds has been extensively studied in the literature. Based on their input representation, these detectors can be categorized as those operating directly on the points~\cite{shi2019pointrcnn,qi2019deep,REF:Yang_3DSSD_2020_CVPR,REF:Point-GNN_CVPR2020,misra2021end,li2021lidar}, on a voxelized space~\cite{REF:Vote3Deep_ICRA2017,REF:VotingforVoting_RSS2015,song2016deep,najibi2020dops,yang2018pixor,REF:simony2018complex,zhou2018voxelnet,REF:pointpillars_cvpr2018,REF:PillarNet_ECCV2020,REF:HVNet2020,zheng2021se}, a perspective projection of the scene~\cite{meyer2019lasernet,REF:bewley2020range,fan2021rangedet}, or a combination of these representations~\cite{sun2021rsn,REF:FastPointRCNN_Jiaya_ICCV2019,zhou2020end,REF:SA_SSD_He_2020_CVPR,shi2020pv}. Semi-supervised 3D detection with a smaller labeled training set or under the annotator-in-the-loop setting have also been considered in~\cite{qi2021offboard,caine2021pseudo,yang2021auto4d}. However, unsupervised 3D detection has been mostly unexplored due to the inherent problem complexity. More recently, Tian \etal~\cite{tian2021unsupervised} proposed to use 3D point cloud clues to perform unsupervised 2D detection in images. In contrast, in our paper, we propose a novel method which performs 3D detection of moving objects in an unsupervised manner.

\subsubsection{Scene Flow Estimation.}
\cutparagraphup
Most previous learning-based works for 3D point cloud scene flow estimation were supervised~\cite{liu2019_flownet3d,wang2020flownet3d++,puy2020flot,gu2019hplflownet}. 
More recently, the unsupervised setting has been also studied. \cite{mittal2020just} used self-supervised cycle consistency and nearest-neighbour losses to train a flow prediction network. In contrast, \cite{li2021_nsf} took an inference-time optimization approach and trained a network per scene. We follow \cite{li2021_nsf} to build our scene flow module given its unsupervised nature and relatively better performance. However, our analysis reveals the limitations of this method in handling complex scenes, making its direct adaptation for proposing high-quality auto labels challenging. In our paper, we noticeably improve the performance of this method by proposing novel techniques to better capture the locality constraints of the scene and to reduce its false predictions.

\subsubsection{Unsupervised Object Detection.}
\cutparagraphup

Existing efforts have been concentrated in the image and video domain, mostly evaluated on object-centric datasets or datasets containing only a handful of object instances per frame.
These include statistic-based methods~\cite{sivic2005discovering,russell2006using}, visual similarity-based clustering methods~\cite{grauman2006unsupervised,faktor2013clustering,joulin2010discriminative}, linkage analysis~\cite{kim2009unsupervised} with appearance and geometric consistency~\cite{cho2015unsupervised,vo2019unsupervised,vo2020toward,vo2021large}, visual saliency~\cite{yuan2009discriminative,jerripothula2016cats}, and unsupervised feature learning using deep generative models~\cite{lee2009convolutional,sohn2013learning,radford2015unsupervised,bau2019gandissect}.
In contrast, unsupervised object detection from LiDAR sequences is fairly under-explored~\cite{dewan2016motion,wong2020identifying,tian2021unsupervised,liu2021opening}.
\cite{dewan2016motion,pang2021model} proposed to sequentially update the detections and perform tracking based on motion cues from 3D LiDAR scans.
Cen~\etal~\cite{cen2021open} used predictions of a supervised detector to yield proposals of unknown categories. However, this approach is inapplicable to unsupervised settings and works for unknown categories with close semantics to the known ones.
Wong~\etal~\cite{wong2020identifying} introduced a bottom-up approach to segment both known and unknown object instances by clustering and aggregating LiDAR points in a single frame based on their embedding similarities.
In comparison, our work leverages both motion cues and point locations for clustering, which puts more emphasis on detecting motion coherent objects and is able to generate amodal bounding boxes.

\subsubsection{Shape Registration.}
\cutparagraphup
Shape registration has been an important topic in vision and graphics community for decades, spanning from classical methods including Iterative Closest Point (ICP)~\cite{besl1992method,chen1992object,rusinkiewicz2001efficient,gross2019alignnet} and Structure-from-Motion (SfM)~\cite{triggs1999bundle,agarwal2010bundle,izadi2011kinectfusion,schoenberger2016sfm} to their deep learning variants~\cite{wang2019deep,ummenhofer2017demon,zhou2017unsupervised,tang2018ba,wei2020deepsfm,yan2016perspective,yan2018learning,tulsiani2017multi,tulsiani2018multi,zhu2017rethinking,insafutdinov2018unsupervised,yang2020teaser}.
These methods usually work under the assumption that the object or scene to register is mostly static or at least non-deformable.
In autonomous driving, shape registration has gained increasing attentions where offline processing is required~\cite{engelmann2016joint,engelmann2017samp,najibi2020dops,stutz2018learning,gu2020weakly,wang2020directshape,duggal2022mending}. 
The shape registration outcome can further support downstream applications 
such as offboard auto labeling~\cite{qi2021offboard,zakharov2020autolabeling,yang2021auto4d}, and perception simulation~\cite{manivasagam2020lidarsim,chen2021geosim}.
In this work, we use sequential ICP with motion-inspired initializations to aggregate partial views of objects and produce the auto-labeled bounding boxes.

\subsubsection{Trajectory Prediction.}
\cutparagraphup
The recent introduction of the large-scale trajectory prediction datasets~\cite{ettinger2021large,caesar2020nuscenes,chang2019argoverse,houston2020one}, helped deep learning based methods to demonstrate new state-of-the-art performance. From a problem formulation stand-point, these methods can be categorized into uni-modal and multi-modal. Uni-modal approaches~\cite{casas2018intentnet,djuric2018short,FaF_Luo_2018_CVPR,gao2020vectornet} predict a single trajectory per agent. Multi-modal methods~\cite{multipath_2019,cui2019multimodal,hong2019rules,bansal2018chauffeurnet,zeng2019end,phan2020covernet,ye2021tpcn,liu2021multimodal,yuan2021agentformer} take into account the possibility of having multiple plausible trajectories per agent. However, all these methods rely on fully labeled datasets. Unsupervised or open-set settings, although practically important for autonomous driving, have so far remained unexplored. Our method enables existing behavior prediction models to generalize to all moving objects, without the need for predefining an object taxonomy.
\cutsectionup
\section{Method}
\label{method}
\cutsectiondown

\begin{figure}[!t]
    \centering
    \includegraphics[width=\linewidth]{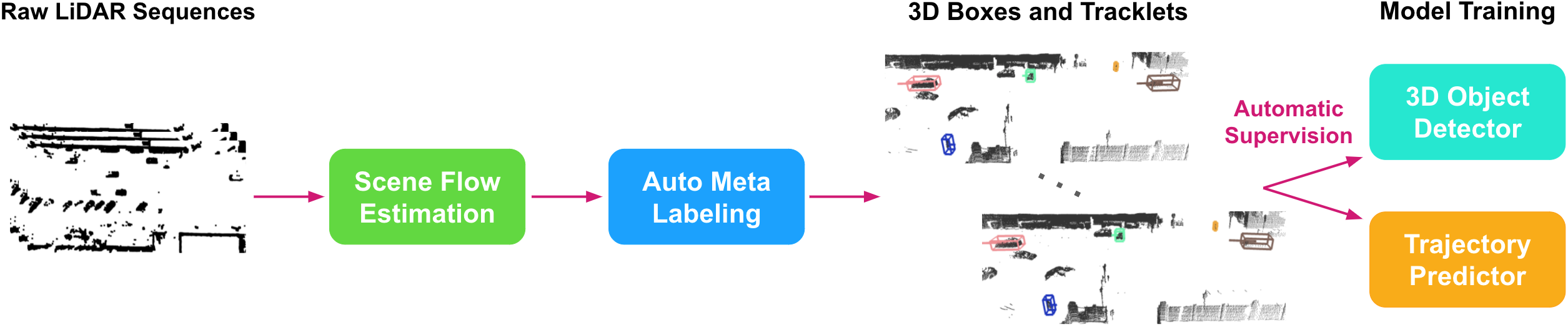}
    \cutcaptionup
    \caption{Proposed framework. Taking as input LiDAR sequences (after ground removal), our approach first reasons about per point motion status (static or dynamic) and predicts accurate scene flow. Based on the motion signal, Auto Meta Labeling clusters points into semantic concepts, connects them across frames and estimates object amodal shapes (3D bounding boxes). The derived amodal boxes and tracklets will serve as automatic supervision to train 3D detection and trajectory prediction models.}
    \label{fig:teaser}
    \cutcaptiondown
\end{figure}

\cref{fig:teaser} illustrates an overview of our proposed method, which primarily relies on motion cues for recognizing moving objects in an unsupervised manner. The pipeline has two main modules: unsupervised scene flow estimation (Sec.~\ref{sec:nsfp_pp}) and Auto Meta Labeling (Sec.~\ref{sec:auto_meta_labeling}).

\cutsubsectionup
\subsection{Neural Scene Flow Prior++}
\label{sec:nsfp_pp}
\subsubsection{Background.}
\cutparagraphup
Many prior works~\cite{Phil_Jund_2022,MotionNet_Wu_2020_CVPR,gu2019hplflownet,liu2019_flownet3d} on scene-flow estimation only considered the supervised scenario where human annotations are available for training. However, these methods cannot generalize well to new environments or to newly seen categories~\cite{li2021_nsf}.  Recently, Li~\etal~\cite{li2021_nsf} propose neural scene flow prior (NSFP), which can learn point-wise 3D flow vectors by solving an optimization problem at run-time without the need of human annotation. Thanks to its unsupervised nature, NSFP can generalize to new environments. It also achieved state-of-the-art performance in 3D scene flow estimation. Still, our study shows that it has notable limitations in handling complex scenes when a mixture of low and high speed objects are present. 
For example, as illustrated in \cref{fig:qualitative_nsfp_nsfp3}, NSFP suffers from underestimating the velocity of moving objects, \ie, false negative flows over pedestrians and inaccurate estimation of fast-moving vehicles. It also introduces excessive false positive flows over static objects (e.g., buildings). We hypothesize that such issues are due to the fact that NSFP applies global optimization to the entire point cloud and the highly diverse velocities of different objects set contradictory learning targets for the network to learn properly.

\subsubsection{Overview.}
\begin{figure}[!t]
    \centering
    \includegraphics[width=1\linewidth]{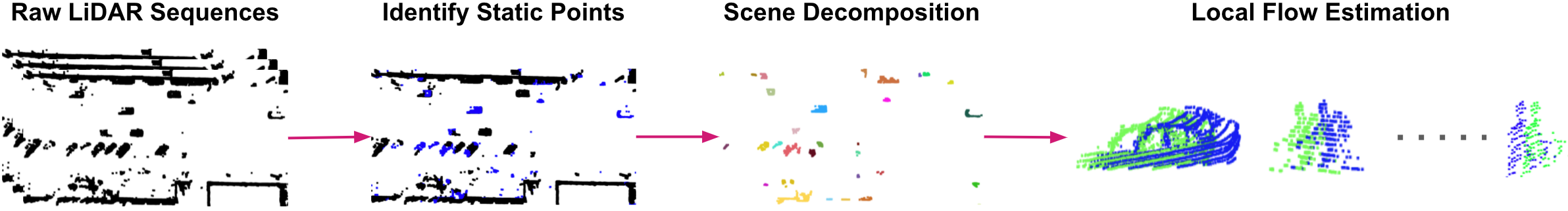}
    \cutcaptionup
    \caption{Proposed NSFP++. Taking as input raw LiDAR sequences (after ground removal), our approach first reasons about the motion status of each point, decomposes the scene into connected components and predicts local flows accurately for each semantically meaningful component.}
    \label{fig:scene_flow}
    \cutcaptiondown
\end{figure}

\cutparagraphup
Our goal is to realize an unsupervised 3D scene flow estimation algorithm that can adapt to various driving scenarios. Here, we present our neural scene flow prior++ (NSFP++) method. As illustrated in \cref{fig:scene_flow}, our method features three key innovations: 1) robustly identifying static points; 2) divide-and-conquer strategy to handle different objects by decomposing a scene into semantically meaningful connected components and targetedly estimating local flow for each of the them; 3) flow consistency among points in each component.
 
\subsubsection{Problem Formulation.}
\cutparagraphup
Let $\mathbf{S}_t \in \mathbb{R}^{N_1 \times 3}$ and $\mathbf{S}_{t+1} \in \mathbb{R}^{N_2 \times 3}$ be two sets of points captured by the LiDAR sensor of an autonomous vehicle at time $t$ and time $t+1$, where $N_1$ and $N_2$ denote the number of points in each set. We denote $\mathbf{F}_t \in \mathbb{R}^{N_1 \times 3}$ as the scene flow, a set of flow vectors corresponding to each point in $\mathbf{S}_t$. Given a point $\mathbf{p} \in \mathbf{S}_t$, we define $\mathbf{f} \in \mathbf{F}_t$ be the corresponding flow vector such that $\hat{\mathbf{p}} = \mathbf{p} + \mathbf{f}$ represents the future position of $\mathbf{p}$ at $t + 1$. Typically, points in $\mathbf{S}_t$ and $\mathbf{S}_{t+1}$ have no correspondence and $N_1$ differs from $N_2$. 

As in Li~\etal~\cite{li2021_nsf}, we model the flow vector $\mathbf{f} = h(\mathbf{p}; \mathbf{\Theta})$ as the output of a neural network $h$, containing a set of learnable parameters as $\mathbf{\Theta}$. To estimate $\mathbf{F}_t$, we solve for $\mathbf{\Theta}$ by minimizing the following objective function:

\begin{small}
\begin{equation}
    \mathbf{\Theta}^*, \mathbf{\Theta}_{\textrm{bwd}}^* = \arg\min\limits_{\mathbf{\Theta}, \mathbf{\Theta}_{\textrm{bwd}}} \sum\limits_{\mathbf{p} \in \mathbf{S}_t} \mathcal{L}(\mathbf{p} + \mathbf{f}, \mathbf{S}_{t+1}) + \sum\limits_{\hat{\mathbf{p}} \in \hat{\mathbf{S}}_t} \mathcal{L}(\hat{\mathbf{p}} + \mathbf{f}_{\textrm{bwd}}, \mathbf{S}_{t})%
\end{equation}
\end{small}
where $\mathbf{f} = h(\mathbf{p}; \mathbf{\Theta})$ is the forward flow, $\mathbf{f}_{\textrm{bwd}} = h(\hat{\mathbf{p}}; \mathbf{\Theta}_{\textrm{bwd}})$ is the backward flow, $\hat{\mathbf{S}}_t$ is the set of predicted future positions for points in $\mathbf{S}_t$ and $\mathcal{L}$ is Chamfer distance function. Here we have the forward and backward flow models share the same network architecture but parameterized by $\mathbf{\Theta}$ and $\mathbf{\Theta}_{\textrm{bwd}}$ respectively. The model parameters, $\mathbf{\Theta}$ and $\mathbf{\Theta}_{\textrm{bwd}}$ are initialized and optimized for each time stamp $t$. Although we only take the forward flow into the next-step processing, learning the flows bidirectionally help improve the scene flow quality~\cite{li2021_nsf,liu2019_flownet3d}. 

\begin{figure}[!t]
    \centering
    \includegraphics[width=0.85\linewidth]{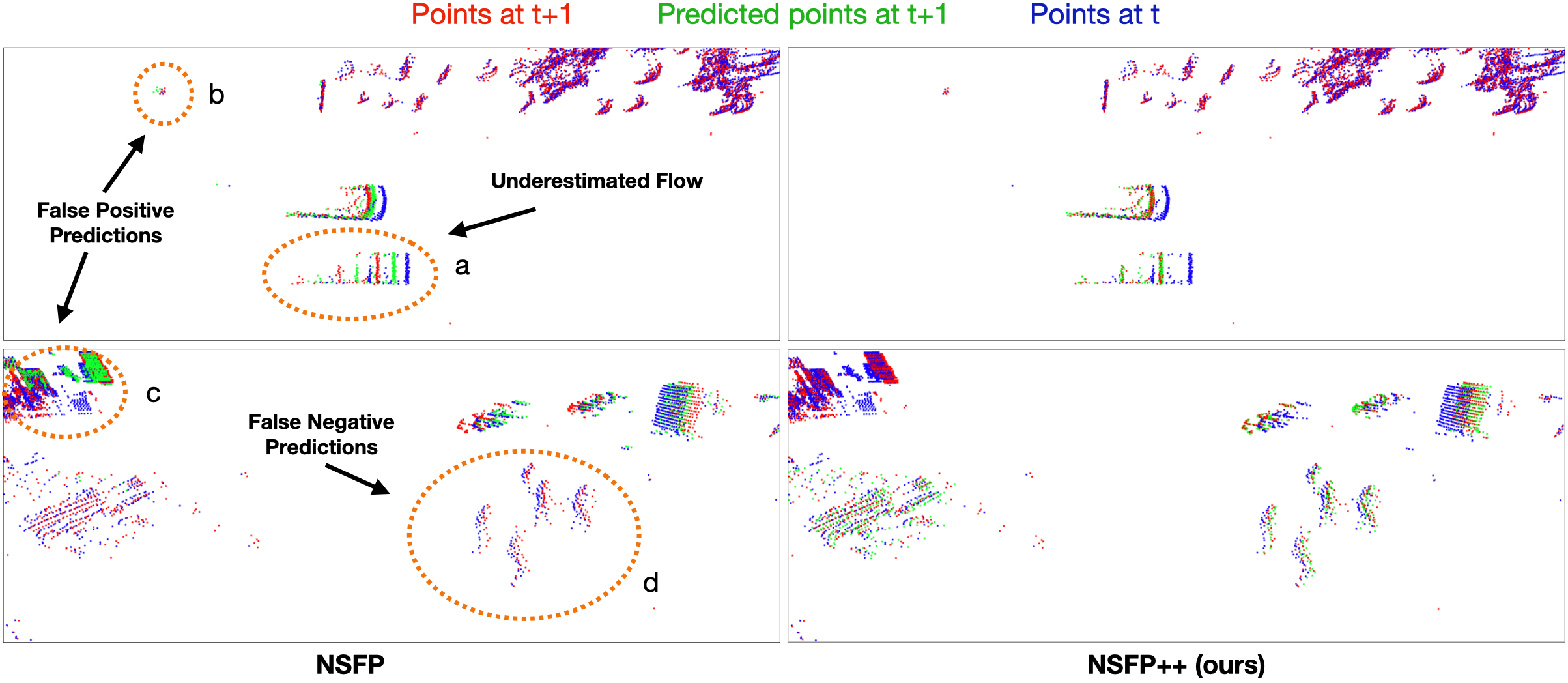}
    \cutcaptionup
    \caption{\small Flow quality comparison between NSFP~\cite{li2021_nsf} and our NSFP++ over the Waymo Open Dataset. Dashed circles in orange color highlight the major shortcomings suffered by NSFP, \textit{i.e.,} (a) underestimated flow for a fast-moving vehicle, (b)(c) false positive predictions at the background and (d) false negative predictions at pedestrians with subtle motion. In contrast, our NSFP++ generates accurate predictions in all these cases.}
    \label{fig:qualitative_nsfp_nsfp3}
    \cutcaptiondown
\end{figure}
 
\subsubsection{Identifying Static Points.}
\cutparagraphup
Since our focus is moving objects, we start by strategically removing static points to reduce computational complexity and benefit scene flow estimation.
In autonomous driving datasets, one large body of static points is ground. Ground is usually captured as a flat surface for which predicting local motion is not possible due to the aperture problem. We follow \cite{li2021_nsf,liu2019_flownet3d} and remove ground points prior to motion estimation. This is achieved by a RANSAC-based algorithm in which a parameterized close-to-horizontal plane is fitted to the points and points in its vicinity are marked as static. However, ground is not the only static part of the scene and unsupervised flow predictions in these static regions (\eg walls, buildings, trees, \etc) introduce noise, reducing the quality of our final auto labels. 
As a result, we further propose to identify more static regions in the scene prior to scene flow estimation. This is achieved by comparing the Chamfer distance between the points in the current frame with those in earlier frames. We mark points as static if the computed Chamfer distance is less than a threshold. We set a small threshold to have a high precision in this step (\ie 20 cm/s in our experiments).

\subsubsection{Estimate Local Flow via Scene Decomposition.}
\cutparagraphup
Inspired by the fact that objects in outdoor scenes are often well-separated after detecting and isolating the ground points, we propose to further decompose the dynamic part of the scene into connected components. This strategy allows us to solve for local flows for each cluster targetedly, which can greatly improve the accuracy of flow estimation for various traffic participants, e.g., vehicles, pedestrians, cyclists, travelling at highly different velocities. 
\cref{fig:scene_flow} gives an overview of our method.

More precisely, given the identified static points, we split the point sets as $\mathbf{S}_t = \mathbf{S}_t^s \cup \mathbf{S}_t^d$ and $\mathbf{S}_{t+1} = \mathbf{S}_{t+1}^s \cup \mathbf{S}_{t+1}^d$, where $\mathbf{S}_t^s$ and $\mathbf{S}_{t+1}^s$ contain static points while $\mathbf{S}_t^d$ and $\mathbf{S}_{t+1}^d$ store dynamic points. This separation, not only helps decompose the scene into semantically meaningful connected components, but also substantially reduces false positive flow predictions on static objects. We then further break down the dynamic points into $\mathbf{S}_t^d = \bigcup_{i=1}^{K} \mathbf{C}_t^i$, where $\mathbf{C}_t^i \in \mathbb{R}^{m_i \times 3}$ is one disjoint cluster of $m_i$ points (the number of clusters $K$ can vary as the scene changes). In the rest of this section, we omit index $i$ for brevity and let $\mathbf{C}_t$ to represent one of the clusters. For every $\mathbf{C}_t \subseteq \mathbf{S}_t^d$ at time $t$, we solve for model parameters to derive local flows, by minimizing the objective function as:
\begin{small}
\begin{eqnarray}
    \mathbf{\Theta}^*, \mathbf{\Theta}_{\textrm{bwd}}^* & = & \arg\min\limits_{\mathbf{\Theta}, \mathbf{\Theta}_{\textrm{bwd}}} 
    \sum\limits_{\mathbf{p} \in \mathbf{C}_t \subseteq \mathbf{S}_t^d}
    \mathcal{L}(\mathbf{p} + \mathbf{f}, \mathbf{C}_{t+1}) +
    \sum\limits_{\hat{\mathbf{p}} \in \hat{\mathbf{C}}_t \subseteq \hat{\mathbf{S}}_t^d}
    \mathcal{L}(\hat{\mathbf{p}} + \mathbf{f}_{\textrm{bwd}}, \mathbf{C}_{t}) \nonumber \\ 
    & + & \frac{\alpha}{|\mathbf{C}_t|} \sum\limits_{\substack{\mathbf{f}_i, \mathbf{f}_j \in \mathbf{F}_{\mathbf{C}_t} \\ i \neq j} } \| \mathbf{f}_i  - \mathbf{f}_j\|_2^2
\end{eqnarray}
\end{small}

\noindent
where the last term is the newly introduced local consistency regularizer with $\alpha$ set to 0.1, $\mathbf{F}_{\mathbf{C}_t}$ consists of flow vectors for each point in $\mathbf{C}_t$, $\hat{\mathbf{S}}_t^d$ contains predicted future positions of all points residing in $\mathbf{S}_t^d$, $\hat{\mathbf{C}}_{t}$ is a subset of $\hat{\mathbf{S}}_t^d$ only storing future positions of points in $\mathbf{C}_t \subseteq \mathbf{S}_t^d$ and $\mathbf{C}_{t+1}$ is a subset of $\mathbf{S}_{t+1}^d$, derived based on box query within a neighborhood of $\mathbf{C}_t$. Next we will present our box query strategy: expansion with pruning.

\begin{figure}[!t]
\centering
\begin{minipage}{.45\textwidth}
  \centering
  \includegraphics[width=0.9\linewidth,height=0.545\linewidth]{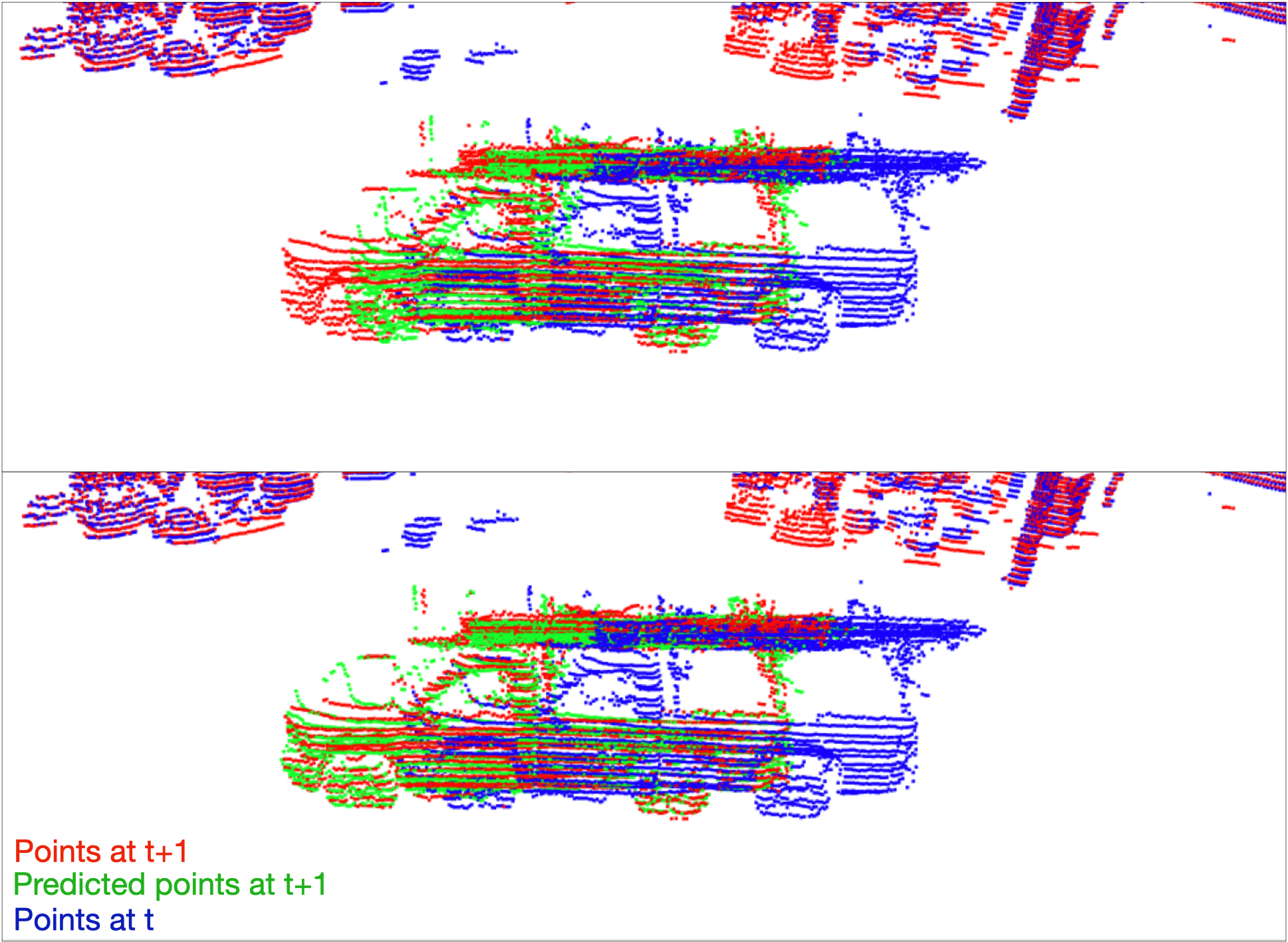}
  \cutcaptionup
  \caption{\small Illustration of the effectiveness of box query with expansion in more accurately estimating flow over the object shape. Top and bottom are without and with expansion.}
  \label{fig:box_query_expansion}
\end{minipage} \hspace{1em}
\begin{minipage}{.46\textwidth}
  \centering
  \includegraphics[width=0.9\linewidth,height=0.525\linewidth]{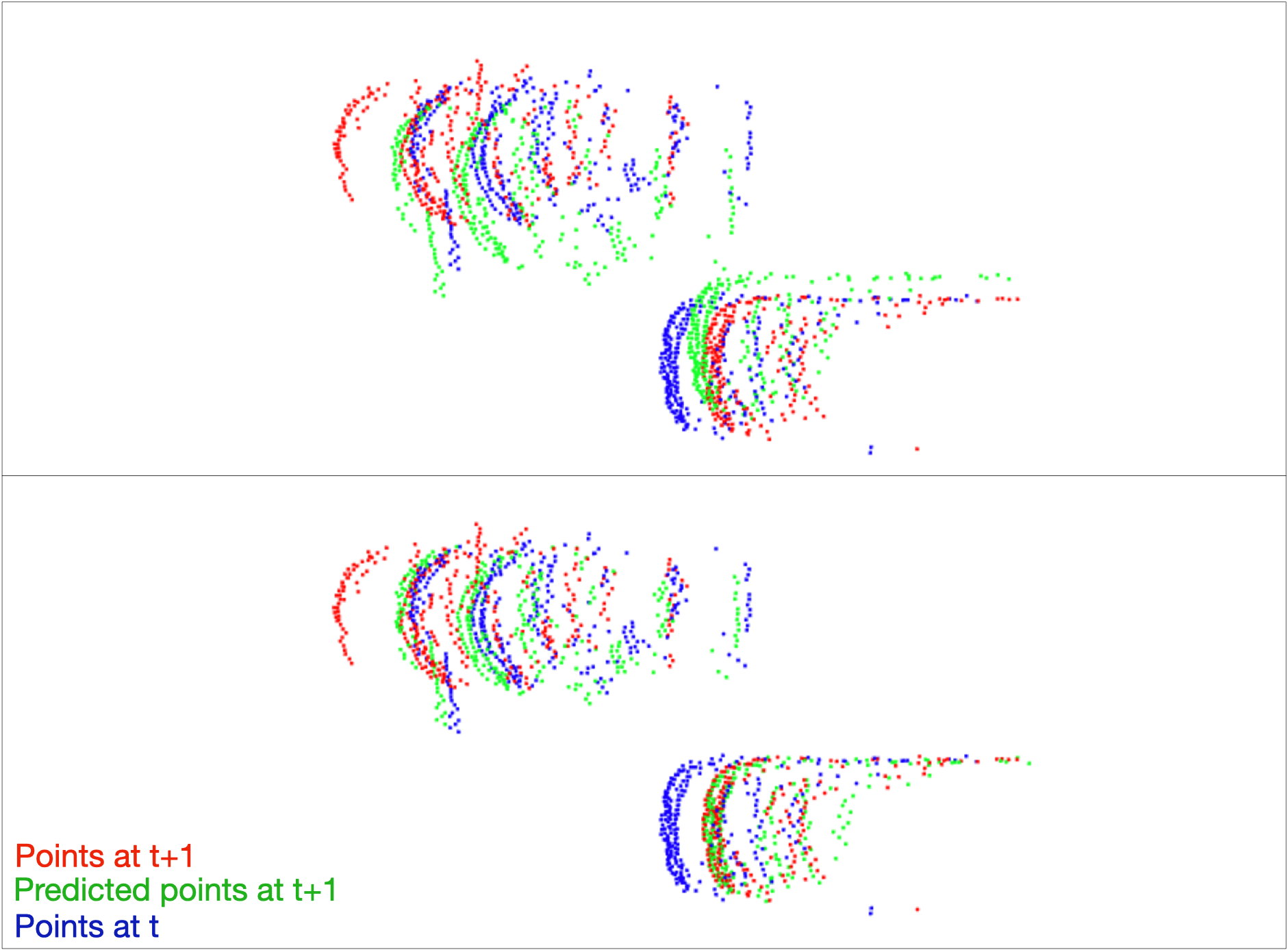}
  \cutcaptionup
  \caption{\small Illustration of the effectiveness of box query followed by pruning in preserving accurately the local flow for nearby objects. Top and bottom are without and with pruning.}
  \label{fig:box_query_pruning}
\end{minipage}

\cutcaptiondown
\end{figure}

\subsubsection{Box Query Strategy.}
\cutparagraphup
Considering that some objects (vehicles) may move at a high speed, we need to expand the field of view to find match points in the next frame. Given a cluster $\mathbf{C}_t$, we find the axis-aligned (along X and Y axes) bounding box tightly covering $\mathbf{C}_t$, in the bird's eye view (BEV). The box is represented as $\mathbf{b} = [x_{\textrm{min}}, y_{\textrm{min}}, x_{\textrm{max}}, y_{\textrm{max}}]$. Note that fast-moving objects, \eg, vehicles, can travel multiple meters between two LiDAR scans. To satisfactorily capture the points of such objects at time $t+1$, we propose to expand the box query with axis-aligned buffer distances $\delta_x$, $\delta_y$ and use $\mathbf{b}^\prime = [x_{\textrm{min}} - \delta_x, y_{\textrm{min}} - \delta_y, x_{\textrm{max}} + \delta_x, y_{\textrm{max}} + \delta_y]$ to retrieve points from $\mathbf{S}_{t+1}^d$, resulting in $\mathbf{C}_{t+1}$. We set the buffer distances according to the aspect ratio of the box $\mathbf{b}$, \textit{i.e.,} $\frac{\delta_y}{\delta_x} = \frac{y_{\textrm{max}} - y_{\textrm{min}}}{x_{\textrm{max}} - x_{\textrm{min}}}$. We empirically set $\max \{{\delta_x}, {\delta_y}\} = 2.5\textrm{m}$. \cref{fig:box_query_expansion} illustrates that expanding box query captures the full shape of a fast-moving truck, resulting in accurate prediction of the future position of the entire object point cloud (\textit{i.e.,} predicted future positions align nicely with the next frame).  

In crowded areas of the scene, retrieved points with $\mathbf{b}^\prime$ may include irrelevant points into the optimization process, causing flow to drift erroneously. See \cref{fig:box_query_pruning} as an example, where two vehicles are moving fast and close to each other. Box query with $\mathbf{b}^\prime$ can include points from the other vehicle and lead to flow drifting. To address this challenge, we propose to prune retrieved points based on the statistics of $\mathbf{C}_{t}$. Formally, let $\mathbf{\Omega}$ be the set of retrieved points by $\mathbf{b}^\prime$ from $\mathbf{S}_{t+1}^d$. We select $n = \min\{|\mathbf{\Omega}|, |\mathbf{C}_{t}|\}$ nearest points from $\mathbf{\Omega}$ with respect to the first moment of $\mathbf{C}_{t}$ and store them in set $\mathbf{C}_{t+1} \in \mathbb{R}^{n \times 3}$. The effectiveness of pruning in keeping relevant points and thus preserving local flow is shown in \cref{fig:box_query_pruning}.

\cutsubsectionup
\subsection{Auto Meta Labeling}
\cutsubsectiondown
\label{sec:auto_meta_labeling}
\begin{figure}[!t]
    \centering
    \includegraphics[width=0.95\linewidth]{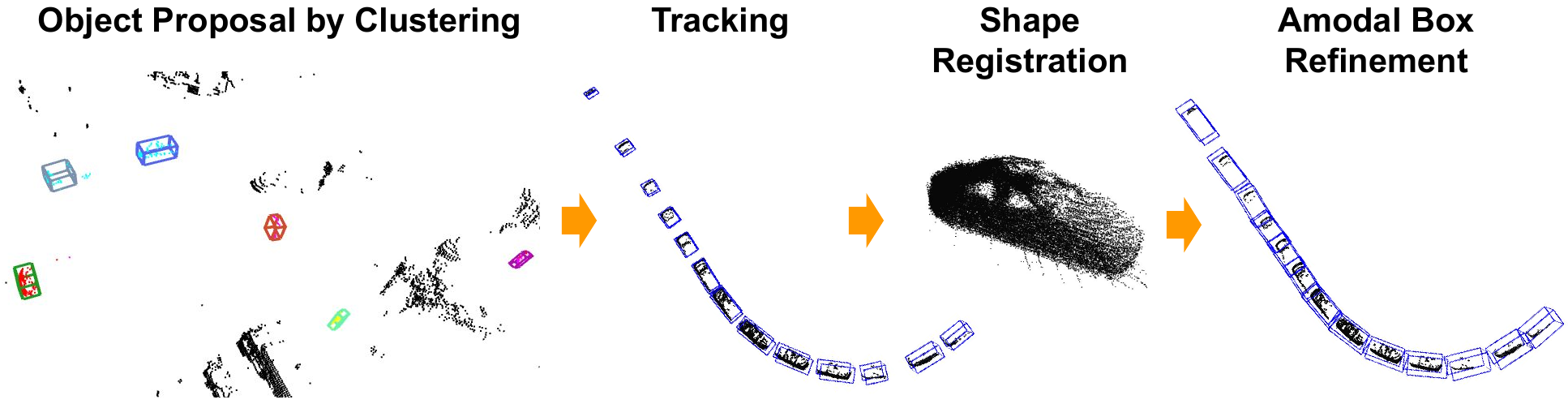}
    \cutcaptionup
    \caption{Auto Meta Labeling pipeline. Given point locations and scene flows on each scene, our Auto Meta Labeling pipeline first proposes objects by spatio-temporal clustering, connects visible bounding boxes of proposals into tracks, then performs shape registration on each track to obtain 3D amodal bounding boxes on each scene.}
    \label{fig:aml_pipeline}
    \cutcaptiondown
\end{figure}

With the motion signals provided by the unsupervised scene flow module, we are able to generate 3D proposals for moving objects without any manual labels. We propose an Auto Meta Labeling pipeline, which takes point clouds and scene flows as inputs and generates high quality 3D auto labels (\cref{fig:aml_pipeline}). The Auto Meta Labeling pipeline has four components: (a) object proposal by clustering, which leverages spatio-temporal information to cluster points into visible boxes (tight boxes covering visible points), forming the concept of objects in each scene; (b) tracking, which connects visible boxes of objects across frames into tracklets; (c) shape registration, which aggregates points of each track to complete the shape for the object; (d) amodal box refinement, which transforms visible boxes into amodal boxes. See supplementary materials for implementation details.

\subsubsection{Object Proposal by Clustering.}\label{sec:clustering}
\cutparagraphup

\begin{figure}[!t]
    \centering
    \includegraphics[width=1\linewidth]{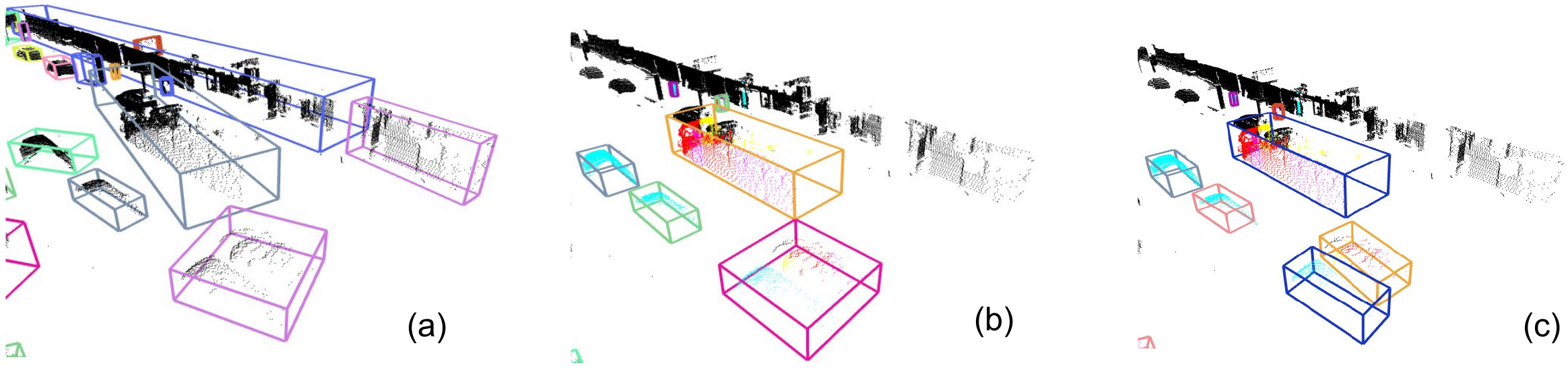}
    \cutcaptionup
    \caption{\small{Comparison between object proposals by different clustering approaches. Points are colored by scene flow magnitudes and directions. Dark for static points. (a) Clustering by location only. (b) Filter by flow magnitude and then cluster based on location (c) Filter by flow and cluster based on both location and motion (ours).}}
    \label{fig:clustering}
    \cuthalfcaptiondown
\end{figure}

On each scene, given the point cloud locations $\mathbf{S} = \{\mathbf{p}_n \mid \mathbf{p}_n \in \mathbb{R}^3 \}_{n=1}^N$ and the corresponding point-wise scene flows $\mathbf{F} = \{\mathbf{f}_n \mid \mathbf{f}_n \in \mathbb{R}^3 \}_{n=1}^N$, the clustering module segments points into subsets where each subset represents an object proposal. We further compute a bounding box of each subset as an object representation. Traditional clustering methods on point cloud often consider 3D point locations $\mathbf{S}$ as the only feature. In the autonomous driving data, with a large portion of points belonging to the background, such methods generate many irrelevant clusters (\cref{fig:clustering}{\color{red}a}). As we focus on moving objects, we leverage the motion signals to reduce false positives. Hence, a clustering method based on both point locations and scene flows is desired.

One simple yet effective strategy can be filtering point cloud by scene flows before object proposal: we only keep points with a flow magnitude larger than a threshold. We then apply the \dbscan~\cite{ester1996density} clustering algorithm on the filtered point sets. This filtering can largely reduce the false positives (\cref{fig:clustering}{\color{red}b}). 

\begin{algorithm}[!t]
\resizebox{0.95\linewidth}{!}{%
\begin{minipage}{\textwidth}
\caption{\small{Object proposal by spatio-temporal clustering on each scene.\label{alg:clustering}}}
\textbf{Input:} point locations $\mathbf{S} = \{\mathbf{p}_n\}_{n=1}^{N}$; point-wise scene flows $\mathbf{F} = \{\mathbf{f_n}\}_{n=1}^{N}$ \\
\textbf{Hyperparams:} neighborhood thresholds $\epsilon_p, \epsilon_f$; minimum flow magnitude $|\mathbf{f}|_{min}$ \\
\textbf{Output:} 3D bounding boxes $\mathbf{B}_{vis} = \{\mathbf{b}_k\}_{k=1}^{M_{pf}}$ of the visible parts of proposed objects
\begin{algorithmic}
\Function{FlowBasedClustering}{$\mathbf{S}$, $\mathbf{F}$; $\epsilon_{p}$, $\epsilon_{f}$, $|\mathbf{f}|_{min}$}:
\State $\mathbf{S}', \mathbf{F}' \leftarrow \text{FilterByFlowMagnitude}(\mathbf{S}, \mathbf{F}; |\mathbf{f}|_{min})$ 
\State $\mathbf{C_p} \leftarrow \text{DBSCAN}(\mathbf{S}'; \epsilon_{p})$ \Comment{$\mathbf{C_p} = \{\mathbf{c}_i\}_{i=1}^{M_p}$, point sets clustered by locations}
\State $\mathbf{C_f} \leftarrow \text{DBSCAN}(\mathbf{F}'; \epsilon_{f})$ \Comment{$\mathbf{C_f} = \{\mathbf{c}_j\}_{j=1}^{M_f}$, point sets clustered by flows}
\For{$\mathbf{c}_i$ in $\mathbf{C_p}$} 
    \For{$\mathbf{c}_j$ in $\mathbf{C_f}$}
        \State $\mathbf{c}_k \leftarrow \mathbf{c}_i \cap \mathbf{c}_j$
        \State $\mathbf{\bar{f}}_k \leftarrow \text{Average}(\{\mathbf{f}_l \mid \forall l : \mathbf{p}_l \in \mathbf{c}_{k}\})$
        \State $\mathbf{b}_k \leftarrow \text{MinAreaBBoxAlongDirection}(\mathbf{c}_k, \mathbf{\bar{f}}_k)$
    \EndFor
\EndFor
\Return $\{\mathbf{b}_k\}_{k=1}^{M_{pf}}$
\EndFunction
\end{algorithmic}
\end{minipage}%
\cutcaptiondown
}
\end{algorithm}

However, there is still a common case where the aforementioned
approach cannot handle well: close-by objects
tend to be under-segmented into a single cluster. 
To solve this issue, we propose clustering by both spatial locations and scene flows (Algorithm \ref{alg:clustering}). After removing points with flow magnitudes smaller than a threshold $|\mathbf{f}|_{min}$, we obtain the filtered point locations $\mathbf{S'}$ and point-wise scene flows $\mathbf{F'}$. 
Then we apply \dbscan~to $\mathbf{S'}$ and $\mathbf{F'}$ separately, resulting in two sets of clusters.
Based on its location and motion, a point may fall into different subsets based on these two clusterings. We then intersect the subsets obtained by the location-based and the flow-based clusterings to formulate the final clusters. In this way, two points are clustered together only if they are close with respect to both their location and motion (\cref{fig:clustering}{\color{red}c}).

Having the cluster label for each point, we form the concept of an object via a bounding box covering each cluster. 
Given the partial observation of objects within a single frame, we only generate boxes tightly covering the visible part in this stage, $\mathbf{B}_{vis} = \{\mathbf{b}_k\}$.
Without object semantics, we use motion information to decide the heading of each box. We compute the average flow $\mathbf{\bar{f}}_k$ of each cluster $\mathbf{c}_k$. Then we find the 7 DoF bounding box $\mathbf{b}_k$ surrounding $\mathbf{c}_k$ which has the minimum area on the $xy$-plane along the chosen heading direction parallel to $\mathbf{\bar{f}}_k$.

\subsubsection{Multi-Object Tracking.}
\cutparagraphup
The tracking module connects visible boxes $\mathbf{B}_{vis}$ into object tracks. Following the tracking-by-detection paradigm ~\cite{weng2019baseline,qi2021offboard}, we use $\mathbf{B}_{vis}$ for data associations and Kalman filter for state updates.
However, rather than relying on the Kalman filter to estimate object speeds, our tracking module leverages our estimated scene flows in the associations.
In each step of the association, we advance previously tracked boxes using scene flows and match the advanced boxes with those in the next frame. 

\cutsubsectionup
\subsubsection{Shape Registration and Amodal Box Refinement.}
\begin{algorithm}[!t]
\resizebox{0.95\linewidth}{!}{%
\begin{minipage}{\textwidth}
\caption{\small{Sequential shape registration and box refinement.\label{alg:registration}}}
\textbf{Input:} An object track with point locations $\{\mathbf{X}_l\}_{l=1}^L$, bounding boxes $\{\mathbf{b}_l\}_{l=1}^L$, headings $\{\theta_l\}_{l=1}^L$. All in world coordinate system. \\
\textbf{Output:} Refined boxes $\{\mathbf{b}'_l\}_{l=1}^L$.
\begin{algorithmic}
\Function{ShapeRegistrationAndBoxRefinement}{$\{\mathbf{X}_l\}_{l=1}^L$, $\{\mathbf{b}_l\}_{l=1}^L$, $\{\theta_l\}_{l=1}^L$}:
\State $\mathbf{X}'_l = \mathbf{X}_l - \mathbf{\bar{X}}_l, \forall l \in \{1, ..., L\}$ \Comment{Normalize points to object-centered}
\State $\mathbf{X}'_{tgt} \leftarrow \mathbf{X}'_{\hat{i}} : \hat{i} = \text{argmax}_i{|\mathbf{X}'_i|}$ \Comment{Init target as the most dense point cloud}
\State $I = \{\hat{i}+1, \hat{i}+2, ..., L, \hat{i}-1, \hat{i}-2, ..., 1\}$ \Comment{Shape registration ordering}
\For{$i$ in $I$}
    \For{$\mathbf{T}_j$ in $\text{SearchGrid}(\mathbf{b}_t)$}
        \State $\mathcal{T}_{\text{init}} \leftarrow [\mathbf{R}_{\theta_{tgt} - \theta_i} \mid \mathbf{T}_j]$ 
        \State $\mathbf{X}'_{tgt, j}, \mathcal{T}_{i \rightarrow tgt, j}, \epsilon_j \leftarrow \text{ICP}(\mathbf{X}'_i, \mathbf{X}'_{tgt}, \mathcal{T}_{\text{init}})$
    \EndFor
    \State $\mathbf{X}'_{tgt}, \mathcal{T}_{i \rightarrow tgt} \leftarrow \mathbf{X}'_{tgt, \hat{j}}, \mathcal{T}_{i \rightarrow tgt, \hat{j}} : \hat{j} = \text{argmin}_j \epsilon_j$ \Comment{Registration w/ least error}
\EndFor
\State $\mathbf{b}'_{tgt} = \text{MinAreaBBoxAlongDirection}(\mathbf{X}'_{tgt} + \mathbf{\bar{X}}_{tgt}, \theta_{tgt})$
\For{$i$ in $I$}
    \State $\mathbf{b}'_i = \text{Transform}(\mathbf{b}'_{tgt}, \mathcal{T}^{-1}_{i \rightarrow tgt})$
\EndFor
\Return $\{\mathbf{b}'_l\}_{l=1}^{L}$
\EndFunction
\end{algorithmic}
\end{minipage}%
\cutcaptiondown
}
\end{algorithm}

In the unsupervised setting, human annotations of object shapes are unavailable. It is hard to infer the amodal shapes of occluded objects purely based on sensor data from one timestamp. However, the observed views of an object often change across time as the autonomous driving car or the object moves. This enables temporal data aggregation to achieve more complete amodal perception of each object.

For temporal aggregation, we propose a shape registration method built upon sequentially applying ICP~\cite{besl1992method,chen1992object,rusinkiewicz2001efficient} (Algorithm \ref{alg:registration}).
ICP performance is sensitive to the transformation initialization. In clustering, we have obtained the headings $\{\theta_l\}_{l=1}^L$ of all visible boxes in each track. The difference in headings of each source and target point set constructs a rotation initialization $R_{\theta_{tgt} - \theta_{src}}$ for ICP. 

In autonomous driving scenarios, shape registration among a sequence of observations poses special challenges: (a) objects are moving with large displacements in the world coordinate system; (b) many observations of objects are very sparse due to their far distance from the sensor and/or heavy occlusions.
These two challenges make it hard to register points from different frames.
To tackle this problem, we search in a grid to obtain the best translation for aligning the source (from frame A) and target (from frame B) point sets. The grid, or the search range, is defined by the size of the target frame bounding box. We initialize the translation $\mathbf{T_j}$ corresponding to different grid points and find the best registration results out of them.

Sequentially, partial views of an object in a track are aggregated into a more complete point set, whose size is often close to amodal boxes. We then compute a bounding box around the target point set similar to the last step in object proposal. During registration, we have estimated the transformation from each source point set to the target, and we can propagate the target bounding box back to each scene by inversing each transformation matrix. Finally, we obtain 3D amodal bounding boxes of detected objects.
\cutsectionup
\section{Experiments}
\label{sec:experiments}
\cutsectiondown

We evaluate our framework using the challenging Waymo Open Dataset (WOD)~\cite{sun2019scalability}, as it provides a large collection of LiDAR sequences with 3D labels for each frame (we only use labels for evaluation unless noted otherwise).
In our experiments, objects with speed $>$ 1m/s are regarded moving.
Hyperparameters and ablation studies are presented in the supplementary material. 

\cutsubsectionup
\subsection{Scene Flow}
\cutsubsectiondown

\subsubsection{Metrics.}
We employ the widely adopted metrics as \cite{li2021_nsf,wu2020pointpwc}, which are 3D end-point error (EPE3D) computed as the mean L2 distance between the prediction and the ground truth for all points; Acc$_5$ denoting the percentage of points with EPE3D $<$ 5cm or relative error $<$ 5\%; Acc$_{10}$ denoting the percentage of points with EPE3D $<$ 10cm or relative error $<$ 10\%; and $\theta$, the mean angle error between predictions and ground truths. In addition, we evaluate our approach based on fine grained speed breakdowns. We assign each point to one speed class (\eg, 0 - 3m/s, 3 - 6m/s, \etc) and employ the Intersection-over-Union (IoU) metric to measure the performance in terms of class-wise IoU and mean IoU. IoU is computed as $\frac{\textrm{TP}}{\textrm{TP} + \textrm{FP} + \textrm{FN}}$, same as in 3D semantic segmentation~\cite{caesar2020nuscenes}.

\subsubsection{Results.}
\cutparagraphup
We evaluate our NSFP++ over all frames of the WOD~\cite{sun2019scalability} validation set and compare it with the previous state-of-the-art scene flow estimator, NSFP~\cite{li2021_nsf}. Following~\cite{Phil_Jund_2022}, we use the provided vehicle pose to compensate for the ego motion, such that our metrics is independent from the autonomous vehicle motion and can better reflect the flow quality on the moving objects.
\cref{fig:qualitative_nsfp_nsfp3} visualizes the improvement of the proposed NSFP++ compared to NSFP. Our approach accurately predicts flows for both high- and low-speed objects (a, d). In addition, NSFP++ not only is highly reliable in detecting the subtle motion of vulnerable road users (d) but can also robustly distinguish all moving objects from the static background (b, c). Finally, our approach outperforms NSFP substantially across all quantitative metrics, as listed in \cref{tab:flow_aggregated_quantitative_wod} and \cref{tab:flow_speed_breakdown_wod}.

\begin{table}[t]
  \centering
 \cutcaptionup
    \caption{Comparison of scene flow methods on the WOD validation set.}
    \label{tab:flow_aggregated_quantitative_wod}
    \resizebox{0.7\linewidth}{!}{%
    \begin{tabular}{c|ccccc}
    \toprule
    Method & \quad EPE3D (m) $\downarrow$ \quad & \quad Acc$_{5}$ (\%) $\uparrow$ \quad & \quad Acc$_{10}$ (\%) $\uparrow$ \quad & \quad $\theta$ (rad) $\downarrow$ \quad \\
    \midrule
    NSFP~\cite{li2021_nsf} & 0.455 & 23.65 & 43.06 & 0.9190 \\

    \midrule
    NSFP++ (ours) & \textbf{0.017} & \textbf{95.05} & \textbf{96.45} & \textbf{0.4737} \\

    \bottomrule
    \end{tabular}%
    }
\cutcaptiondown
\end{table}%

\begin{table}[t]
  \centering
    \caption{Comparison of scene flow methods on the WOD validation set, with speed breakdowns.}
     \label{tab:flow_speed_breakdown_wod}
    \resizebox{0.7\linewidth}{!}{%
    \begin{tabular}{c|cccccc|c}
    \toprule
    \multirow{2}{*}{Method} & \multicolumn{6}{c|}{IoU per Speed Breakdown (m/s)} & \multirow{2}[2]{*}{mIOU} \\
    & \quad 0 - 3 \quad & \quad 3 - 6 \quad & \quad 6 - 9 \quad & \quad 9 - 12 \quad & \quad 12 - 15 \quad & \quad 15+ \quad &  \\
    \midrule
    NSFP~\cite{li2021_nsf} & 0.657 & 0.152 & 0.216 & 0.166 & 0.130 & 0.140 & 0.244 \\
    
    \midrule
    NSFP++ (ours) &  \textbf{0.989} & \textbf{0.474} & \textbf{0.522} & \textbf{0.479} & \textbf{0.442} & \textbf{0.608} & \textbf{0.586} \\
    \bottomrule
    \end{tabular}%
    }
  \cutcaptiondown
\end{table}

\cutsubsectionup
\subsection{Unsupervised 3D Object Detection}\label{sec:exp_det}
\cutsubsectiondown
\begin{table}[t]
  \centering
  \footnotesize
   \bgroup
    \def\tabcolsep{10pt}
    \cutcaptionup
    \caption{Comparisons between 3D detectors trained with autolabels generated by AML with supervised flow and unsupervised flow.}
      \label{tab:pointpollar_results}
    \resizebox{0.85\linewidth}{!}{
    \begin{tabular}{c|c|cc|cc}
    \toprule
     \multirow{2}{*}{Method} & \multirow{2}{*}{Supervision} & \multicolumn{2}{c|}{3D mAP}  & \multicolumn{2}{c}{2D mAP}   \\ 
     & & L1 & L2 & L1 & L2 \\ \midrule
     Sup Flow~\cite{Phil_Jund_2022} + Clustering & \multirow{2}{*}{Supervised} & 30.8 & 29.7 & 42.7 & 41.2 \\ 
     Sup Flow~\cite{Phil_Jund_2022} + AML & & 49.9 & 48.0 & 56.8 & 54.8   \\
     \midrule
     No flow + Clustering & \multirow{4}{*}{Unsupervised} & 4.7 & 4.5 & 5.8 & 5.6 \\ 
     No flow + AML & & 9.6 & 9.4 & 11.0 & 10.8 \\ 
     Unsup Flow + Clustering & & 30.4 & 29.2 & 36.7 & 35.3 \\ 
     Unsup Flow + AML & & \textbf{42.1} & \textbf{40.4} & \textbf{49.1} & \textbf{47.4}   \\
     \bottomrule
    \end{tabular}
    }
    \egroup

  \cutcaptiondown
  \end{table}

Our method aims at generating auto labels for training downstream autonomous driving tasks in a fully unsupervised manner. 3D object detection is a core component in autonomous driving systems. In this section, we evaluate the effectiveness of our unsupervised AML pseudo labels by training a 3D object detector. We adopt the PointPillars~\cite{REF:pointpillars_cvpr2018} detector for our experiments. All models are trained and evaluated on WOD~\cite{sun2019scalability} training and validation sets. Since there is no category information during training, we use a single-class detector to detect any moving objects. We train and evaluate the detectors on a 100m x 40m rectangular region around the ego vehicle to reflect the egocentric importance of the predictions \cite{deng2021revisiting}. We set a 3D $\textrm{IoU}$ of 0.4 during evaluation to count for the large variation in size of the class-agnostic moving objects, e.g., vehicles, pedestrians, cyclists. We employ a top-performing flow model~\cite{Phil_Jund_2022} as the supervised counterpart to our unsupervised flow model NSFP++.

\cref{tab:pointpollar_results} compares performance of detectors trained with auto labels generated by our pipelines and several baselines. The first two rows show detection results when a fully supervised flow model~\cite{Phil_Jund_2022} (flow supervision derived from human box labels) is deployed for generating the auto labels. The first row represents a baseline where our hybrid clustering method is used to form the auto labels based on motion cues~\cite{dewan2016motion}. The second row shows the performance when the same supervised flow predictions are used in combination with our AML pipeline. Clearly, our AML pipeline greatly outperforms the clustering baseline, verifying the high-quality auto labels generated by our method. The last four rows consider the unsupervised setting. \textit{No flow + Clustering} is a baseline where \dbscan~is applied to the point locations to form the auto labels. \textit{No flow + AML} is our pipeline when purely relying on a regular tracker without using any flow information. \textit{Unsup Flow + Clustering} uses our proposed hybrid clustering technique on the outputs of our NSFP++ scene flow estimator without connecting with our AML. \textit{Usup Flow + AML} is our full unsupervised pipeline. Notably, not only does it outperforms other unsupervised baselines by a large margin, but it also achieves a comparable performance with the supervised \textit{Sup Flow + AML} counterpart. Moreover, comparing it with other unsupervised baselines by removing parts of our pipeline validates the importance of all components in our design (please see the supplementary for more ablations). Most importantly, our approach is a fully unsupervised 3D pipeline, capable of detecting moving objects in the open-set environment. This new feature is cost efficient and safety critical for the autonomous vehicle to reliably detect arbitrary moving objects, removing the need of human annotation and the constraint of predefined taxonomy.

\cutsectionup
\subsection{Open-set 3D Object Detection}
\cutsectiondown
In this section, we turn our attention to the open-set setting where only a subset of categories are annotated. Since there is no public 3D dataset designed for this purpose, we perform experiments in a leave-one-out manner on WOD~\cite{sun2019scalability}. WOD has three categories, namely vehicle, pedestrian, and cyclist. Considering the similar appearances and safety requirements, we combine pedestrian and cyclist into a larger category called VRU (vulnerable road user), resulting in a data size comparable with the vehicle category. We then assume to only have access to human annotations for one of the two categories, leaving the other one out for our auto meta label pipeline to pseudo label.

\cref{tab:openset_det} shows the results. The first two rows show the performance of a fully supervised point pillars detector. As expected, when the detector is trained on one of the categories, it can not generalize to the other. In the last two rows, when human annotations are not available, we rely on our auto labels to fill in for the unknown category. When no vehicle label is available, our pipeline helps the detector to generalize and consequently improves the mAP from 48.8 to 77.1. Although generalizing to VRUs without any human labels is a more challenging scenario, our pipeline still improves the mAP by a noticeable margin, showing its effectiveness in the open-set settings.

\begin{table}[ht]
\centering
\begin{small}
\cutcaptionup
\caption{Open-set 3D detection results.}
\label{tab:openset_det}
\resizebox{0.75\linewidth}{!}{%
\begin{tabular}{c|c|c|c|c|c}
\toprule
\multirow{2}{*}{} & \multicolumn{2}{c|}{Human Labeled}    & \multirow{2}{*}{Vehicle 3D AP} & \multirow{2}{*}{VRU 3D AP} & \multirow{2}{*}{3D mAP} \\ \cline{2-3}
                  &  Vehicle & VRU &                   &                   &                   \\ \midrule
\multirow{2}{*}{Supervised Method} & \checkmark &  & 97.5& 0.0& 48.8\\ \cline{2-6} 
                  &  & \checkmark & 0.0 &88.7 & 44.4 \\ \midrule
\multirow{2}{*}{Ours (Supervised + AML)} & \checkmark &  & 97.5 & 20.8 & \textbf{59.2}\\ \cline{2-6} 
                  &  & \checkmark & 65.4 & 88.7& \textbf{77.1}\\
\bottomrule
\end{tabular}
}
\end{small}

\cutcaptiondown
\end{table}

\cutsubsectionup
\subsection{Open-set Trajectory Prediction}
\cutsubsectiondown
For trajectory prediction, we have extracted road graph information for a subset of WOD (consisting of 625 training and 172 validation sequences). 
We use those WOD run segments with road graph information for our trajectory prediction experiments.
Following~\cite{ettinger2021large}, a trajectory prediction model is required to forecast the future positions for surrounding agents for 8 seconds into the future, based on the observation of 1 second history. We use the MultiPath++~\cite{MultiPathPP_2021} model for our study. The model predicts 6 different trajectories for each object and a probability for each trajectory. To evaluate the impact of open-set moving objects on the behavior prediction task, we train models using perception labels derived via different strategies as the ground truth data and then evaluate the behavior prediction metrics of the trained models on a manually labeled validation set. We use the minADE and minFDE metrics as described in~\cite{ettinger2021large}.

\cref{tab:openset_bp} reports the trajectory prediction results. While the supervised method achieves a reasonable result when the vehicle class is labeled, its performance is poor when trained only on the VRU class. This is expected, as the motion learned from slow vehicles can be generalized to VRUs to some extent, but predicting the trajectory of the fast moving vehicles is out of reach for a model trained on only VRUs. The last two rows show the performance of the same model when AML is deployed for auto-labeling the missing category. Consistent with our observation in 3D detection, our method can bridge the gap in the open-set setting. Namely, our approach significantly remedies the generalization problem from VRUs to vehicles and achieves the best performance when combining human labels of the vehicle class with our auto labels for VRUs.
\begin{table}[ht]
\cutcaptionup
\centering
\caption{Open-set trajectory prediction results.}
\label{tab:openset_bp}
\resizebox{0.6\linewidth}{!}{%
\begin{tabular}{c|c|c|c|c}
\toprule
\multirow{2}{*}{} & \multicolumn{2}{c|}{Human Labeled}    & \multirow{2}{*}{minADE} & \multirow{2}{*}{minFDE} \\ \cline{2-3}
                  &  Vehicle & VRU &                   &                  \\ \midrule
\multirow{2}{*}{Supervised Method} & \checkmark &  & 2.12 & 5.39\\ \cline{2-5} 
                  &  & \checkmark & 9.53 & 22.31\\ \midrule
\multirow{2}{*}{Ours (Supervised + AML)} & \checkmark &  & \textbf{1.89} & \textbf{4.79} \\ \cline{2-5} 
                  &  & \checkmark & 2.15 & 5.55\\
\bottomrule
\end{tabular}
}

\cutcaptiondown
\end{table}

\cutsectionup
\section{Conclusion}
\label{conclusion}
\cutsectiondown
In this paper, we proposed a novel unsupervised framework for training onboard 3D detection and prediction models to understand open-set moving objects. Extensive experiments show that our unsupervised approach is competitive in regular detection tasks to the counterpart which uses supervised scene flow. With promising results, it demonstrates great potential in enabling perception and prediction systems to handle open-set moving objects. We hope our findings encourage more research toward solving autonomy in an open-set environment.

{\small
\bibliographystyle{splncs04}
\bibliography{ref}
}

\newpage
\appendix
\section*{Appendix}
\section{Implementation Details of Auto Meta Labeling}

The Auto Meta Labeling pipeline has four components: object proposal by clustering, multi-object tracking and amodal box refinement based on shape registration.
In the object proposal step, we use DBSCAN for both clustering by point locations and by scene flows. Both clustering methods use Euclidean distance as the distance metric. The neighborhood thresholds $\epsilon_p$ and $\epsilon_f$ are set to be $1.0$ and $0.1$, respectively. The minimum flow magnitude $|\mathbf{f}|_{min}$ is set to 1m/s, so as to include meaningful motions without introducing too much background noise. Our tracker follows the implementation as in~\cite{qi2021offboard}. We use bird's eye view (BEV) boxes for data association and use Hungarian matching with an IoU threshold of 0.1. In shape registration, we use a constrained ICP~\cite{gross2019alignnet} which limits the rotation to be only round $z$-axis. We have compared the effect of contrained and unconstrained ICP in AML ablation study. The search grid for translation initialization is decided by the target box dimensions on the $xy$-plane, \ie the length $l_{\mathbf{b}_{tgt}}$ and the width $w_{\mathbf{b}_{tgt}}$ of the target bounding box. We enumerate translation initialization $\mathbf{T}_j$ in a $5 \times 5$ grid covering the target bounding box region with a list $\mathcal{T}_x$ of strides as $[-l_{\mathbf{b}_{tgt}} / 2, -l_{\mathbf{b}_{tgt}} / 4, 0, l_{\mathbf{b}_{tgt}} / 4, l_{\mathbf{b}_{tgt}} / 2]$ and a list $\mathcal{T}_y$ of strides $[-w_{\mathbf{b}_{tgt}} / 2, -w_{\mathbf{b}_{tgt}} / 4, 0, w_{\mathbf{b}_{tgt}} / 4, w_{\mathbf{b}_{tgt}} / 2]$. Each computation of ICP outputs an error $\epsilon_j$, which is defined as the mean of the Euclidean distances among matched points between the source and the target point sets.
\section{Ablation Study on Unsupervised Flow Estimation}

In this section we provide additional ablation studies focusing on our unsupervised flow estimation method, NSFP++. 

\subsubsection{Static point removal}
As mentioned in \cref{sec:nsfp_pp}, we apply static point removal prior to unsupervised flow estimation. This step is designed to achieve a high precision to avoid removing dynamic points in the early stages of our pipeline. Here, we compute the precision/recall of this step on the WOD~\cite{sun2019scalability} validation set. We define ground-truth dynamic/static labels based on the available ground-truth bounding boxes~\cite{Phil_Jund_2022}. Dynamic points are defined as those with a ground-truth flow magnitude larger than $|\mathbf{f}|_{min}$, and the remaining points belonging to any ground-truth box 
are assigned to the static class. Our static point removal step has a precision of 97.2\%, and a recall of 62.2\%, validating the high precision of this step in determining the static points.

\begin{table}[t]
\footnotesize
  \centering
    \caption{Ablation study on different components in the proposed local flow estimation. BQ stands for the proposed box query strategy, which contains two steps, the first being expansion and the second being pruning. Local consistency represents the local consistency loss among flow predictions within each point cluster.}
    \resizebox{1.0\linewidth}{!}{
    \begin{tabular}{c|c|c|c|c|c}
    \toprule
    \multicolumn{3}{c|}{Variants of NSFP++} & \multirow{2}{*}{\, EPE3D $\downarrow$ \,} & 
    \multirow{2}{*}{\, $\theta$ (rad) $\downarrow$ \,} & \multirow{2}{*}{\, mIoU $\uparrow$ \,} \\
    BQ w. Expansion & BQ w. Pruning & Local Consistency & & \\
    \midrule
    & & & 0.020 & 0.515 & 0.404 \\
    \checkmark & & & 0.023 & 0.560 & 0.552 \\
    \checkmark & \checkmark & & 0.018 & 0.504 & 0.571 \\
    \checkmark & \checkmark & \checkmark & 0.017 & 0.474 & 0.586 \\
    \bottomrule
    \end{tabular}%
    }
  \label{tab:ablation_study_nsfpp_wod}%
\end{table}%

\begin{table}[t]
\footnotesize
  \centering
    \caption{Flow comparison with the fully supervised model.}
    \begin{tabular}{c|c|c|c}
    \toprule
    Method &  \quad EPE3D (m) $\downarrow$  \quad &  \quad $\theta$ (rad) $\downarrow$ \quad & \quad mIoU $\uparrow$ \\
    \midrule
    Fully Supervised Network & 0.005 & 0.062 & 0.826 \\

    \midrule
    Unsupervised NSFP++ (ours) & 0.017 & 0.474 & 0.586 \\

    \bottomrule
    \end{tabular}%
  \label{tab:supp_unsup_sup_flow}%
  \cutcaptiondown
\end{table}

\subsubsection{Local flow estimation}
We also conduct ablation study to validate the effectiveness of the proposed components in the local flow estimation step, \textit{i.e.,} box query with expansion followed by pruning and local consistency loss. As illustrated in \Cref{tab:ablation_study_nsfpp_wod}, box query with expansion (second row) effectively boosts mIoU from 0.404 to 0.552 but suffers from higher 3D end-point error (EPE3D) and mean angle error ($\theta$), compared to the method without using box query (first row). This is due to the fact that the expanded query region can capture more matching points but at the cost of including irrelevant points. With the proposed pruning scheme (third row), all metrics are significantly improved compared to the previous two rows. Finally, by adding local consistency loss (fourth row), we obtain the best performance across the board.

\subsubsection{Comparison with the fully supervised model}
In this subsection, we compare our unsupervised flow estimation method with  the fully supervised scene flow model used in \cref{sec:experiments}. Table \ref{tab:supp_unsup_sup_flow} shows the comparison. As expected, the supervised model outperforms our unsupervised NSFP++ method which does not use any human annotations. However, as shown in \cref{tab:pointpollar_results}, the AML pipeline can robustly use our unsupervised NSFP++ predictions and eventually achieves comparable results to the counterpart using a supervised flow model on downstream tasks (e.g., L1 mAP of 42.1 for unsupervised \textit{v.s.} 49.9 for supervised in the object detection task).

\section{Ablation Study on Auto Meta Labeling}

\begin{table}[t]
  \centering
  \caption{Comparisons of different variants of components in the AML pipeline. All methods are evaluated on the WOD validation set.}
   \bgroup
    \def\tabcolsep{10pt}
    \resizebox{\linewidth}{!}{%
    \begin{tabular}{c|cc|cc}
    \toprule
     \multirow{2}{*}{AML Variants} & \multicolumn{2}{c|}{3D mAP}  & \multicolumn{2}{c}{2D mAP}   \\ 
     & L1 & L2 & L1 & L2 \\  \midrule
     Filtered by flow + Clustering by position & 25.5 & 24.6 & 32.4 & 31.2 \\ 
     Spatio-temporal clustering & \textbf{30.4} & \textbf{29.2} & \textbf{36.7} & \textbf{35.3} \\
     \midrule
     Regis. w/o init. & 32.2 & 31.0 & 36.6 & 35.3 \\ 
     Regis. w/ R init. by flow heading & 33.2 & 31.9 & 37.4 & 36.0 \\ 
     Regis. w/ T init. by grid search & 35.2 & 33.8 & 39.3 & 37.9 \\ 
     Regis. w/ Unconstrained ICP & 34.3 & 33.0 & 38.5 & 37.1 \\ 
     Regis. w/ RT init. \& constrained ICP~\cite{gross2019alignnet} (\textbf{Full AML}) & \textbf{36.9} & \textbf{35.5} & \textbf{40.5} & \textbf{39.0} \\
     \bottomrule
    \end{tabular}
    }
    \egroup
  \label{tab:aml_ablation}
  \end{table}
  
To examine the design choices in the AML pipeline, we compute the detection metrics on the auto labels generated by our full AML pipeline and several baselines (\Cref{tab:aml_ablation}). Note that the numbers reported in \Cref{tab:aml_ablation} are from evaluation on auto labels, rather than on the predictions by trained detectors. \textit{Filtered by flow + Clustering by position} is a baseline where we generate auto labels only using this clustering method. Compared to our spatial-temporal clustering method described in \cref{alg:clustering}, this baseline does not perform clustering on the estimated flows and as a result it leads to under-segmentation and lower performance.

We also carry out experiments on variants of shape registration. \textit{Regis. w/o init.} is a baseline where we have no initialization when performing constrained ICP. Adding either rotation initialization by flow heading (\textit{Regis. w/ R init. by flow heading}) or translation initialization by grid search (\textit{Regis. w/ T init. by grid search}) improves the quality of auto labels. Another baseline, \textit{Regis. w/ Unconstrained ICP}, is applying both $\mathbf{R}$ and $\mathbf{T}$ initializations but uses an unconstrained ICP such that 3D rotations are allowed when aligning the source and the target point sets. We find that limiting the rotation to be only around $z$-axis generates auto labels with a higher quality. Finally, our full AML (\textit{Regis. w/ RT init. \& constrained ICP}) outperforms all other variants. 
Compared to the 3D detection results in the main paper (see \cref{tab:pointpollar_results}), we find that the object detector achieves higher mAP than the auto labels it is trained on.
The reason is that auto labels by design pursue high recall while contain some false positives in the background due to inaccurate flow or noise in the environment. As these false positive labels do not form a consistent data pattern, the object detector learns to focus only on auto labels with common patterns, such as vehicles and VRUs, and assign high confidence scores to these objects at inference time.

\section{Qualitative Analysis}

\subsection{Auto Meta Labeling and Unsupervised Object Detection}

\begin{figure}
    \centering
    \includegraphics[width=0.95\textwidth]{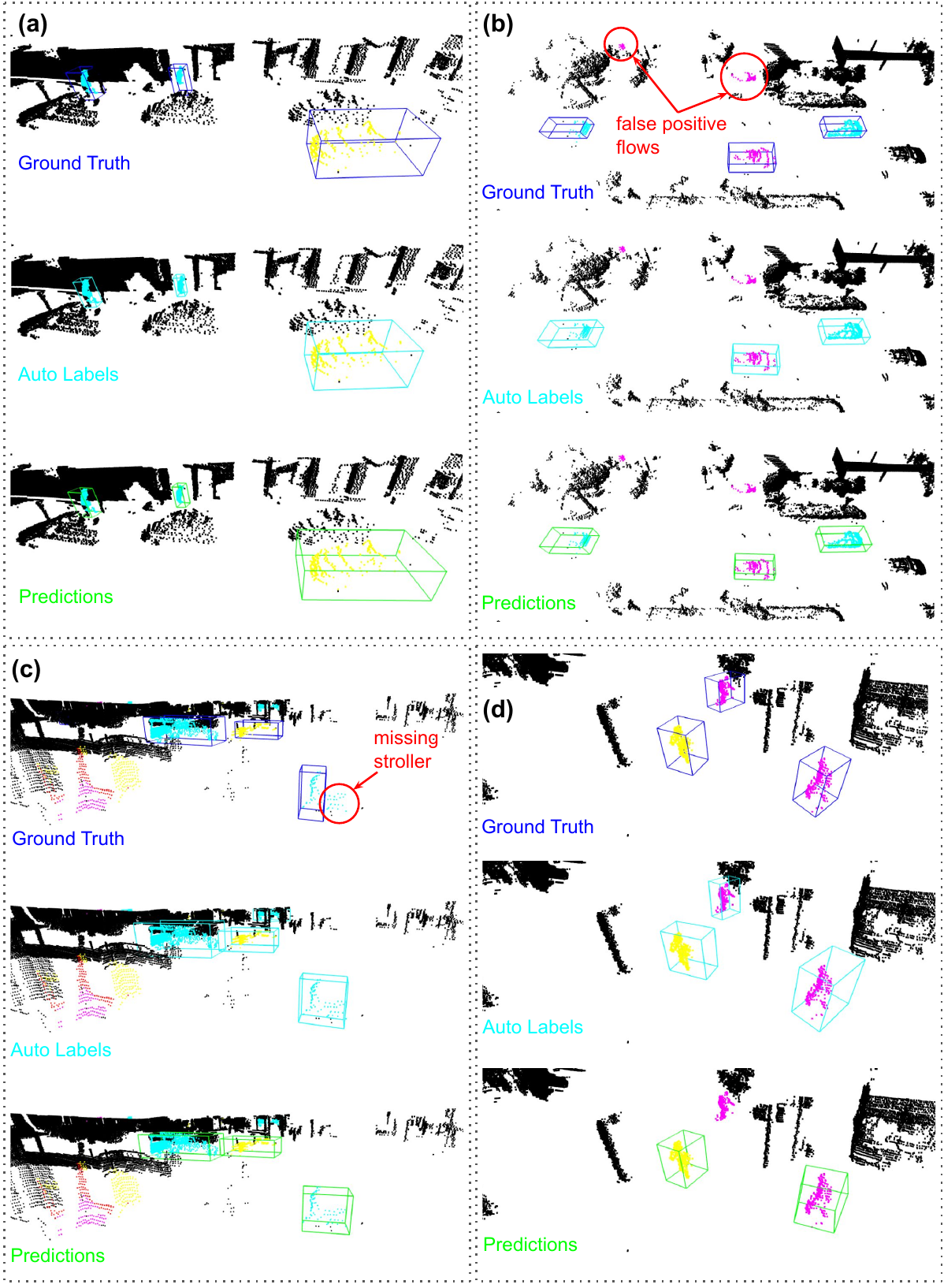}
    \caption{Visualization of auto labels and detection predictions compared with the ground truth of moving objects. Points are colored by flow magnitudes and directions. Dark points are static. (a) The class-agnostic auto labels and unsupervised object detectors capture objects of multiple categories. (b) Although false positive flows occur, AML filters out many of them if they are inconsistent, and the detector learns to ignore these false positive flows. (c) Although the ground truth does not cover categories beyond vehicle, pedestrian, and cyclist, auto labels and our detector can capture open-set moving objects, such as the stroller. (d) An failure case that the detector may not be confident on objects with limited data amount, such as cyclists.}
    \label{fig:al_det_viz}
\end{figure}

\cref{fig:al_det_viz} shows four examples from the WOD validation set comparing ground truth, auto labels and unsupervised object detection results. In our unsupervised setting, both the auto labels and object detectors localize objects in a class-agnostic manner and are not limited by certain categories. In example (a) we show that auto labels and object detectors capture both pedestrians and vehicles.

In example (b), we demonstrate that even though there is \textit{false positive} non-zero flow estimation, in AML we filter out many of these clusters during tracking and post-processing where very short tracks are dropped. The resulting detector has also learned to ignore clusters of false positive flows. This example also shows that both auto labels and object detectors can infer the amodal boxes of some objects with only partial views.

Sometimes the unsupervised flow estimation captures \textit{true positive} motion on points that are beyond the predefined categories in the ground truth. In example (c), a pedestrian is walking with a stroller while \textit{stroller} is not a class included in the ground truth labels and therefore no bounding box is annotated around the stroller. NSFP++ has estimated the flow on the stroller, enabling AML and detectors to localize it. Since the stroller is held by the pedestrian with a similar speed, the clustering by design does not separate them apart. Clearly, it is safety-critical for autonomous vehicles to understand such moving objects in the open-set environment.

Example (d) shows a failure case where the detector could not confidently detect a cyclist. Although the auto labels have captured it, cyclists are less common than pedestrians and vehicles in the training set, which leads to inferior performance. We encourage future work to tackle the data imbalance issue under the unsupervised setting. Another failure pattern is that bounding boxes in auto labels tend to be larger than the actual size, due to the fact that temporal aggregation can include noise points.
More advanced shape registration methods may help reduce noise and we leave it for future work. 

\subsection{Open-set Trajectory Prediction}

\begin{figure}
    \centering
    \includegraphics[width=\textwidth]{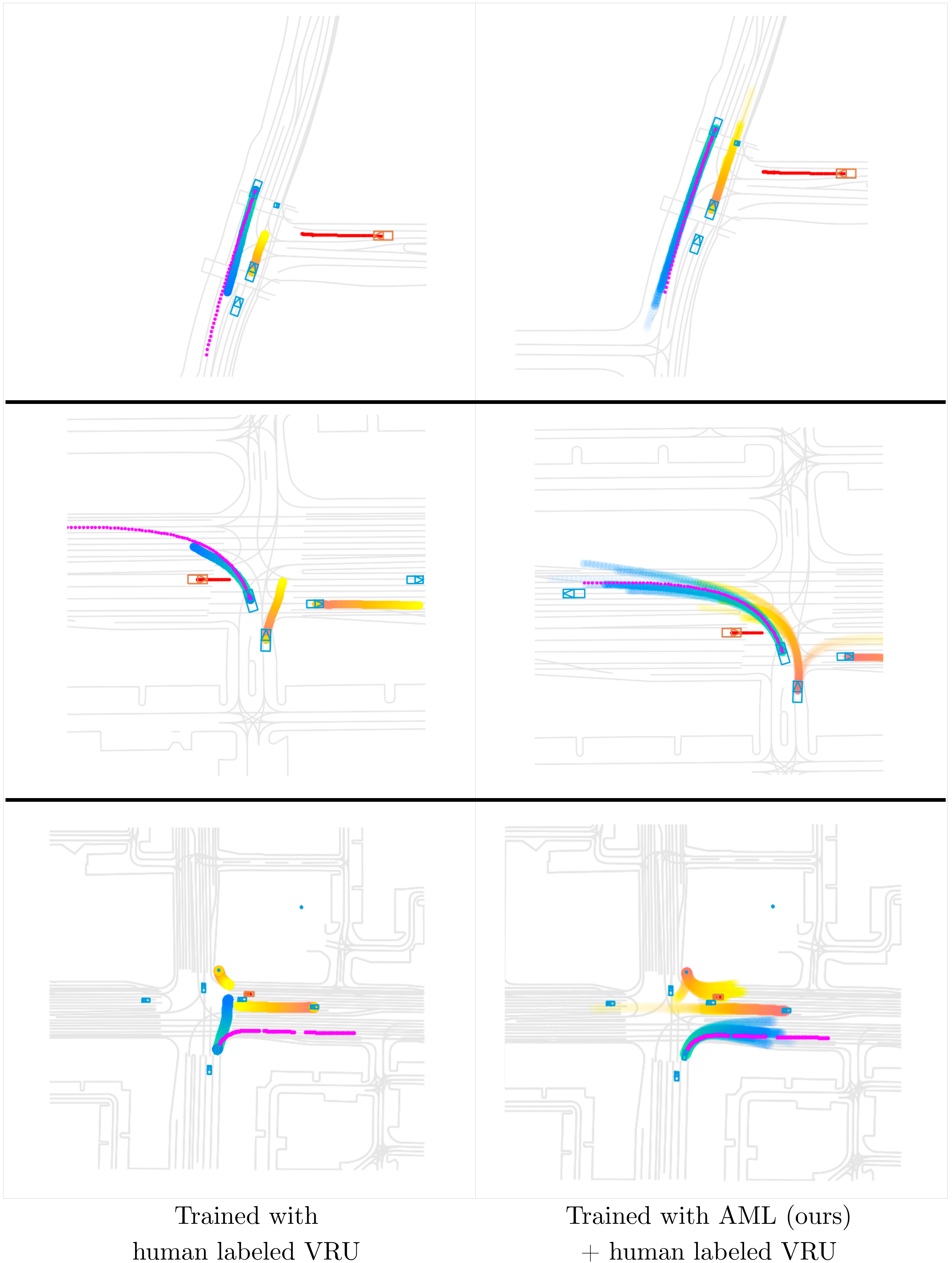}
    \caption{Behavior prediction qualitative analysis. Trajectory predictions on three example scenarios for a model trained with human labeled VRUs \textit{v.s.} a model trained with a combination of human labeled VRUs and our generated autolabels. Red and magenta dotted trajectories represent the ground-truth routes of the autonomous vehicle and agents, respectively. Blue and yellow trajectories are the predictions for the agent of interest and other agents, respectively.}
    \label{fig:bp_vis}
\end{figure}

\begin{figure}
    \centering
    \includegraphics[width=\textwidth]{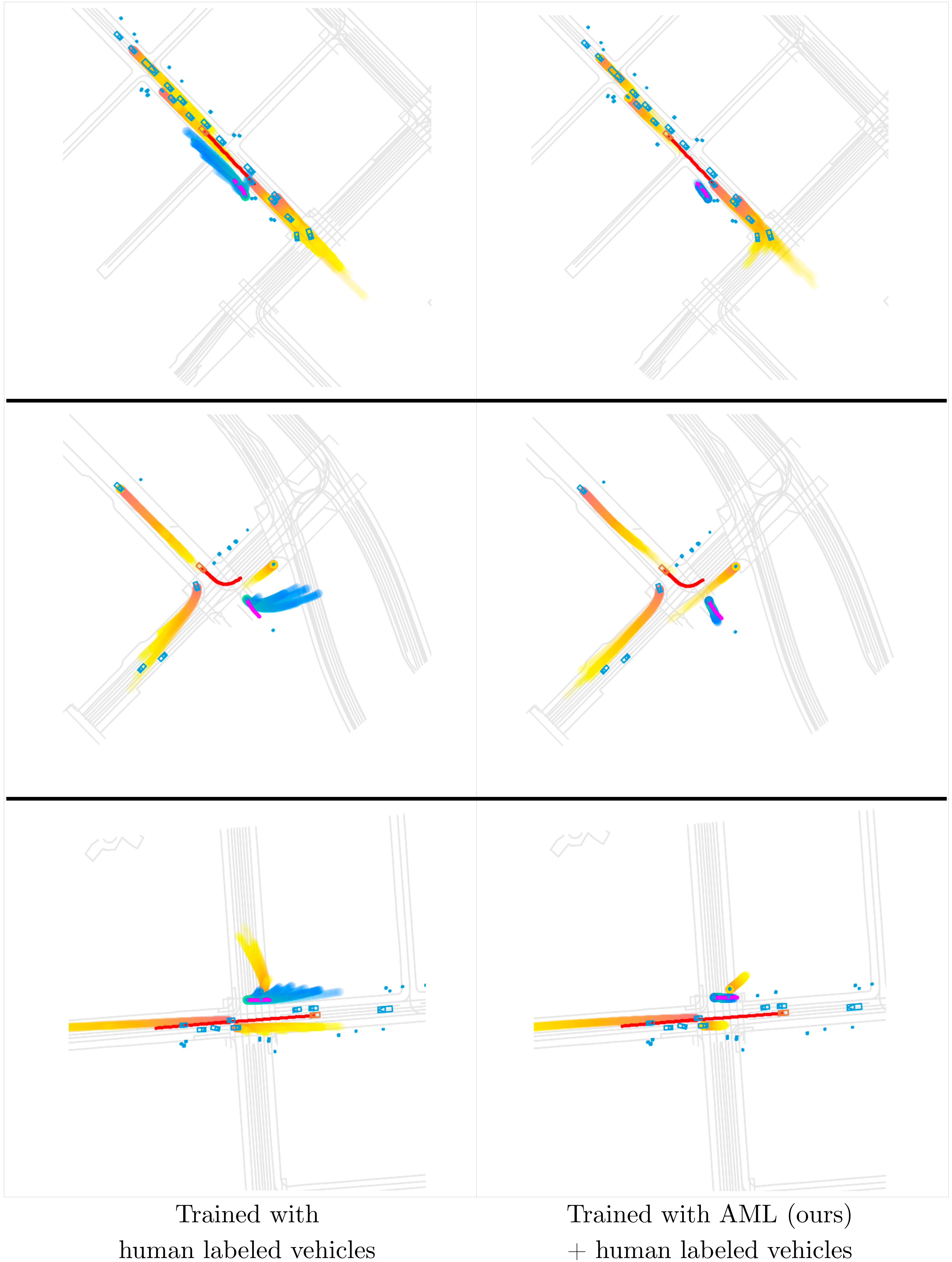}
    \caption{Behavior prediction qualitative analysis. Trajectory predictions on three example scenarios for a model trained with human labeled vehicles \textit{v.s.} a model trained with a combination of human labeled vehicles and our generated autolabels.}
    \label{fig:bp_vis_manual_veh}
\end{figure}

\cref{fig:bp_vis} and \ref{fig:bp_vis_manual_veh} show behavior prediction qualitative results on the validation set of our newly created Anonymized Dataset. For each example scenario, we show the trajectory predictions of two models, \textit{i.e.,} one trained only with a human-labeled category (the first column) and the other one trained with the combination of available human-labels and our AML auto labels for all other moving objects (the second column). The red and magenta trajectories represent the ground-truth routes taken by the autonomous vehicle and by an agent of interest, respectively. The blue and yellow trajectories are the possible predictions for the agent of interest and other agents in the scene. \cref{fig:bp_vis} shows three examples where human labels are available for the VRU category. As can be seen in all three examples, without using our unsupervised auto labels, the model tends to erroneously underestimate the speed (\textit{e.g.} the first row), have difficulty in predicting trajectories consistent with the underlying roadgraph (\textit{e.g.} the second row), and generating dangerous pedestrian-like trajectories along the pedestrian crosswalk (\textit{e.g.} the third row). \cref{fig:bp_vis_manual_veh} shows the results when human labels are available only for the vehicle category. Similarly, when the model is only trained on the human labels (the first column), it cannot generalize well to the VRU class, predicting fast speeds and vehicle-like trajectories for VRUs. However, in both scenarios, adding auto labels (the second columns in \cref{fig:bp_vis} and \ref{fig:bp_vis_manual_veh}) satisfactorily overcomes these errors, showing the effectiveness of our auto labels for training behavior prediction models in the open-set environment.

\section{Failure Analysis}

\begin{figure}[t]
    \centering
    \includegraphics[width=\textwidth]{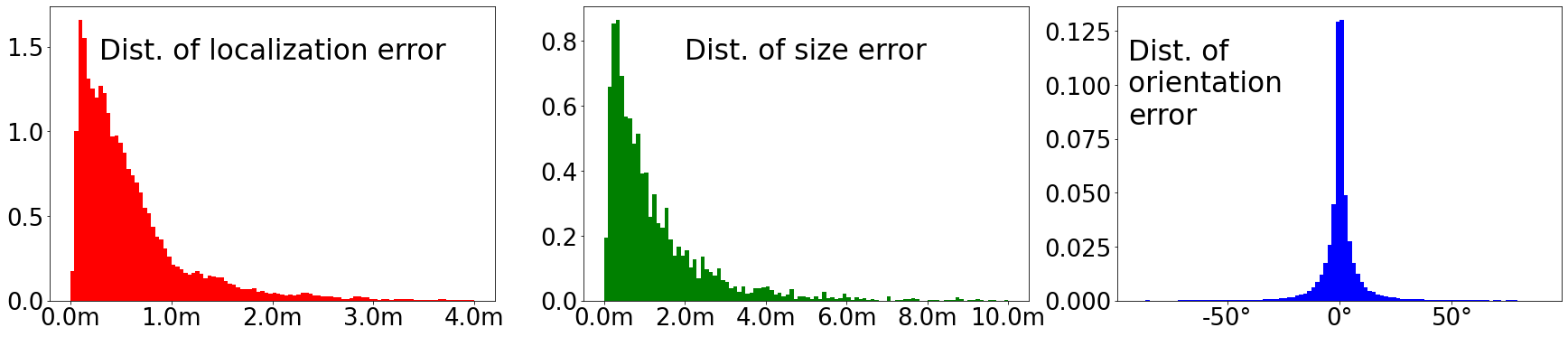}
    \caption{Error distributions. y axis is probability density.}
    \label{fig:failure_analysis}
\end{figure}

In this section, we analyze the factors causing failure cases. Under threshold IoU=0.4, the precision/recall of our auto meta labels is 0.69/0.50. Part of the failure cases come from (1) false positive predictions that do not match any ground truth boxes; (2) false negatives where ground truth boxes are entirely missed. Moreover, there are predicted boxes overlapping with ground truth boxes while their IoUs are lower than the threshold. To have a better understanding, we breakdown 3D bounding box dimensions into three groups: localization (box center $x, y, z$), size (box length $l$, width $w$, height $h$), and orientation (BEV box heading $r$). Then, we summarize the distributions of localization, size, and orientation errors of the generated bounding boxes which overlap with at least one ground truth box (Fig. \ref{fig:failure_analysis}). The errors are computed between each pair of a predicted box and the ground truth box that has the highest IoU with the predicted box.

\subsubsection{Localization.}
The localization error is defined as
\begin{equation}
    \epsilon_{localization} = \sqrt{(x_{pr} - x_{gt})^2 + (y_{pr} - y_{gt})^2 + (z_{pr} - z_{gt})^2}.
\end{equation}
As shown in Fig. \ref{fig:failure_analysis}, most of the localization errors are within 1.0 meter.

\subsubsection{Size.}
The size error is defined as
\begin{equation}
    \epsilon_{size} = \text{max}\{ |l_{pr} - l_{gt}| + |w_{pr} - w_{gt}| + |h_{pr} - h_{gt}| \}.
\end{equation}
Many predictions have relatively high size errors. This is often caused by inclusion of noisy points in the registration step or missing parts of an object if the parts are always invisible throughout the object track.

\subsubsection{Orientation.}
The orientation error is defined as 
\begin{equation}
    \epsilon_{orientation} = r_{pr} - r_{gt}
\end{equation}
The orientation errors are generally small, as the orientation of each object is determined by the direction of the scene flows averaged over all points within the object bounding box. This error distribution verifies the quality of the unsupervised scene flows.

To find out the dominant factors leading to wrong auto meta labels, we construct several baselines by modifying the predictions and measure their label quality. The baselines are as follows:
\begin{enumerate}
    \item (Oracle) GT localization + GT size + GT orientation: we replace the 7D values ($x, y, z, l, w, h, r$) of each predicted box with the values of its best matched ground truth box if any;
    \item Predicted localization + GT size + GT orientation: we replace the ($l, w, h, r$) of each predicted box with the ground truth values. Comparison with the oracle will show the impact of localization errors; 
    \item GT localization + Predicted size + GT orientation: we replace the ($x, y, z, r$) of each predicted box with the ground truth values. Comparison with the oracle will show the impact of size errors; 
    \item GT localization + GT size + Predicted orientation: we replace the ($x, y, z, l, w, h$) of each predicted box with the ground truth values. Comparison with the oracle will show the impact of orientation errors.
\end{enumerate}

We report the 3D mAPH@IoU=0.4 on the above baselines as mAPH additionally reflect the quality of heading prediction. We found that localization and size errors are dominant factors and future work may focus on improving the quality of auto labels on these fronts.

\begin{table}[t]
\centering
\caption{Comparison between an oracle with GT box coordinates and baselines switching localization/size/orientation coordinates into AML predictions in turn. The performance drops show that the localization and size errors are dominant.}
\begin{tabular}{l|c}
\toprule
                                                    & 3D mAPH@IoU=0.4 \\ \midrule
(Oracle) GT localization + GT size + GT orientation & 46.1            \\
Predicted localization + GT size + GT orientation   & 39.7 (-6.4)     \\
GT localization + Predicted size + GT orientation   & 39.7 (-6.4)     \\
GT localization + GT size + Predicted orientation   & 44.5 (-1.6)     \\
\bottomrule
\end{tabular}
\end{table}

\end{document}